%% file: neurips_2022.tex
\documentclass{article}

\usepackage[preprint]{neurips_2022}

\usepackage[utf8]{inputenc} 
\usepackage[T1]{fontenc}    
\usepackage{hyperref}       
\usepackage{url}            
\usepackage{booktabs}       
\usepackage{amsfonts}       
\usepackage{nicefrac}       
\usepackage{microtype}      
\usepackage{xcolor}         
\usepackage{graphicx}
\usepackage{url}
\usepackage{listings}
\usepackage{multirow}
\usepackage{subfigure}

\hypersetup{
    colorlinks=true,
    linkcolor=red,
    citecolor=cyan,
    filecolor=magenta,      
    urlcolor=cyan,
    }

\input{math_commands.tex}

\newcommand{\ie}{\textit{i.e.}}

\title{FaceMAE: Privacy-Preserving Face Recognition \\via Masked Autoencoders}

\author{
  Kai Wang$^{1}$\thanks{Equal contribution. (kai.wang@comp.nus.edu.sg, bo.zhao@ed.ac.uk)},\; Bo Zhao$^{2}$\footnotemark[1],\; Xiangyu Peng$^{1}$,\; Zheng Zhu$^{3}$,\; Jiankang Deng$^{4}$,\;\\ \textbf{Xinchao Wang}$^{1}$,\; \textbf{Hakan Bilen}$^{2}$,\; \textbf{Yang You}$^{1}$\thanks{Corresponding author (youy@comp.nus.edu.sg).}
   \\
  $^{1}$National University of Singapore \quad
  $^{2}$The University of Edinburgh \\
  $^{3}$Tsinghua University \quad
  $^{4}$InsightFace\\
Code: \url{https://github.com/kaiwang960112/FaceMAE}
}

\begin{document}

\maketitle

\begin{abstract}

Face recognition, as one of the most successful applications in artificial intelligence, has been widely used in security, administration, advertising, and healthcare. However, the privacy issues of public face datasets have attracted increasing attention in recent years. Previous works simply mask most areas of faces or synthesize samples using generative models to construct privacy-preserving face datasets, which overlooks the trade-off between privacy protection and data utility. In this paper, we propose a novel framework FaceMAE, where the face privacy and recognition performance are considered simultaneously. Firstly, randomly masked face images are used to train the reconstruction module in FaceMAE. We tailor the instance relation matching (IRM) module to minimize the distribution gap between real faces and FaceMAE reconstructed ones. During the deployment phase, we use trained FaceMAE to reconstruct images from masked faces of unseen identities without extra training. The risk of privacy leakage is measured based on face retrieval between reconstructed and original datasets. Experiments prove that the identities of reconstructed images are difficult to be retrieved. We also perform sufficient privacy-preserving face recognition on several public face datasets (\ie CASIA-WebFace and WebFace260M).  Compared to previous state of the arts, FaceMAE consistently \textbf{reduces at least 50\% error rate} on LFW, CFP-FP and AgeDB.

\end{abstract}

\input{tex/introduction}

\input{tex/relatedwork}

\input{tex/method}
\input{tex/experiments}

\input{tex/conclusion}

\textbf{Acknowledge.} This research is supported by the National Research Foundation, Singapore under its AI Singapore Programme (AISG Award No: AISG2-PhD-2021-08-008), NUS ARTIC Project (ECT-RP2), China Scholarship Council 201806010331, the EPSRC programme grant Visual AI EP/T028572/1, AISG (Award No.: AISG2-RP-2021-023) and MOE Tier 1/Faculty Research Committee Grant (WBS: A-0009440-00-00). We thank Google TFRC for supporting us to get access to the Cloud TPUs. We thank CSCS (Swiss National Supercomputing Centre) for supporting us to get access to the Piz Daint supercomputer. We thank TACC (Texas Advanced Computing Center) for supporting us to get access to the Longhorn supercomputer and the Frontera supercomputer. We thank LuxProvide (Luxembourg national supercomputer HPC organization) for supporting us to get access to the MeluXina supercomputer.

{\small
\bibliographystyle{splncs04}
\bibliography{references}}



\end{document}

%% file: math_commands.tex
\usepackage{amsmath,amsfonts,bm}

\def\eqref#1{equation~\ref{#1}}

\def\1{\bm{1}}

\DeclareMathAlphabet{\mathsfit}{\encodingdefault}{\sfdefault}{m}{sl}
\SetMathAlphabet{\mathsfit}{bold}{\encodingdefault}{\sfdefault}{bx}{n}



%% file: tex/introduction.tex
\section{Introduction}

In the past decade, face recognition has achieved remarkable and continuous progress in improving recognition accuracy \cite{deng2019arcface, wen2016discriminative, an2021partial, zhu2021webface260m, wang2021efficient, yi2014learning, sun2014deep} and has been widely used in daily activities such as online payment and security for identification. 
Advanced face recognition algorithms \cite{deng2019arcface, wen2016discriminative, an2021partial, wang2021efficient, wang2019co, wang2021facex, zeng2020npcface, zhang2019improved, wang2018support, he2010maximum, wu2018light, zhang2016joint, wen2016discriminative} and large-scale public face datasets \cite{zhu2021webface260m, yi2014learning, liu2018large} are two key factors of these progresses.
Nevertheless, collecting and releasing large-scale face datasets raise increasingly more concerns on the privacy leakage of identity membership \cite{hayes2019logan, wu2019privacy} and attribute \cite{mirjalili2020privacynet, bortolato2020learning} of training samples in recent years.
Generating large-scale privacy-preserving face datasets for downstream tasks is urgent and challenging for face recognition community \cite{wu2019privacy, mirjalili2020privacynet, qiu2021synface}. 

This paper focuses on the membership privacy of a public face dataset instead of inferring specific training samples or model parameters \cite{rigaki2020survey, de2020overview}. Specifically, in this scenario, the adversary aims to infer whether target identities are in the training set by retrieving them with query face images when the privacy-preserving face dataset is accessible.
Protecting such membership privacy is crucial for practical applications.  Our goal is to reduce the face retrieval accuracy, in other words, reduce the risk of membership privacy leakage, while keep the informativeness for training deep models on the privacy-preserving face dataset.


\begin{figure}[htp]
    \centering
    \includegraphics[width=\columnwidth]{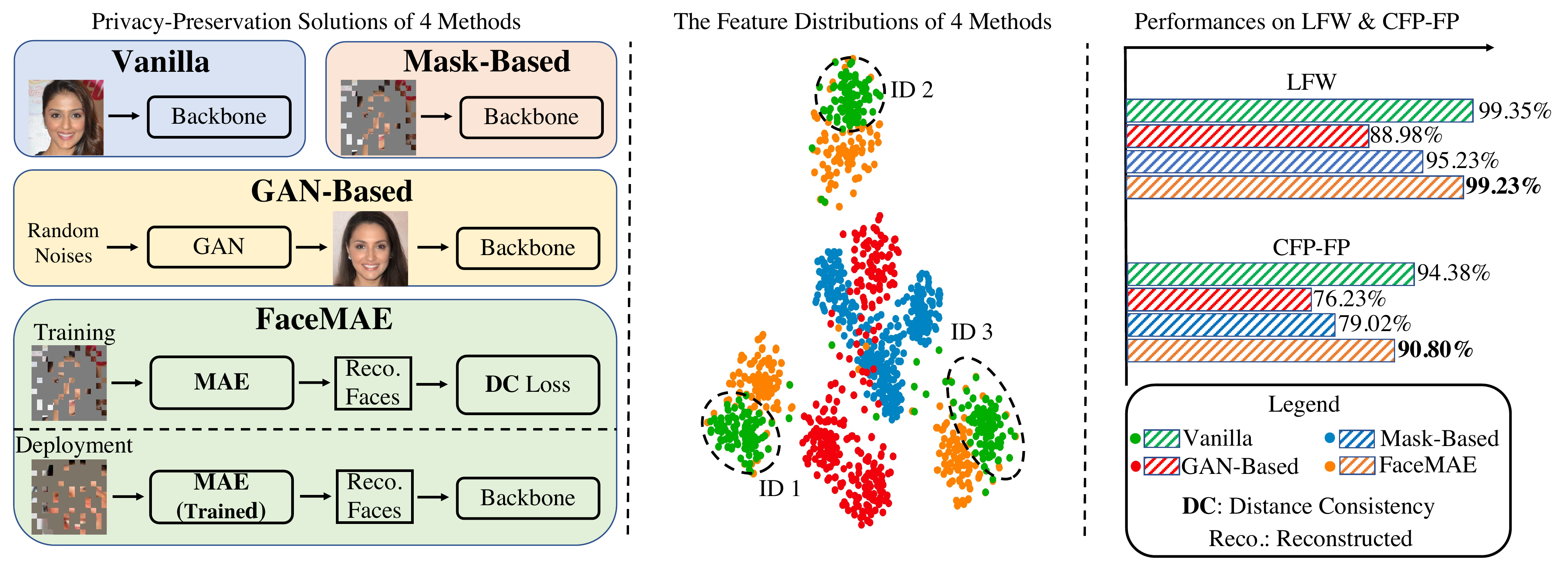}
    \caption{The left part shows four training paradigms of face recognition, namely vanilla, mask-based, GAN-based methods and our FaceMAE. The abbreviation `Reco.' means `Reconstructed', and `DC' means `Distance Consistency'. We also visualize the feature distributions of face images in central part. It is obviously that the face images learned by FaceMAE have more similar distribution with original faces. The right panel shows the performance comparisons of four methods on LFW and CFP-FP datasets. Compared to training on original and generated faces, mask-based and FaceMAE methods need less (25\%) visible information of the whole faces. FaceMAE can capture the original face feature distribution more accurately thus it outperforms GAN-based and mask-based methods.}
    \label{fig:motivation}
\end{figure}

As illustrated in Fig. \ref{fig:motivation}, traditionally distortion such as blurring, noising and masking is applied to face images for reducing the privacy \cite{dufaux2010framework, korshunov2013using}. These naive distortion methods reduce the privacy {as well as} semantics in a face image, thus producing unsatisfying recognition performance.
The recent works try to generate privacy-preserving face dataset by synthesizing identity-ambiguous faces \cite{qiu2021synface}, de-identification faces \cite{wu2019privacy, li2019anonymousnet} or attribute-removed faces \cite{mirjalili2020privacynet, bortolato2020learning} with adapted generative adversarial networks (GANs). 
They can generate as many synthetic faces as required for privacy-preserving face recognition task.
Although these GAN-based methods can generate real-looking faces, the utility of these generated faces for training deep models are not guaranteed. As shown in the central part of Fig. \ref{fig:motivation}, there exists remarkable domain gap between generated and original faces, which leads to poor face recognition performance of GAN-based methods \cite{qiu2021synface}.
Another drawback of GAN based methods is that the generators trained on one dataset cannot be deployed on unseen identities.
In addition, all raw face images of the target dataset are required to train the generative models, which causes extra privacy leakage risks.

For membership privacy, an intuitive sense is that a raw face contains more privacy than the masked one. 
As shown in Fig. \ref{fig:motivation}, the features distribution of masked faces are close to each other, which indicates the masked faces only contains little privacy.
This naturally raises a question: ``Is it feasible to reconstruct informative face images from these masked faces that are good for training deep models?''
In this paper, we propose a novel framework FaceMAE, where the face privacy and recognition performance are considered simultaneously.
Specifically, FaceMAE consists of two stages, named training and deployment.
In the training stage, we adapt masked autoencoders (MAE) \cite{he2021masked} to reconstruct a new dataset from randomly masked face dataset.
Distinct from vanilla MAE, the objective of FaceMAE is to generate the faces those are beneficial for face recognition training. Therefore, we utilize instance relation matching module (IRM) instead of Mean Squared Error loss (MSE-Loss) in vanilla MAE, which aims to minimize the difference between the relation graphs of original and reconstructed faces.
To the best of our knowledge, we are the first to tailor a new optimization object rather than inherit the MSE-Loss for MAE.
Once the training stage is finished, we apply the trained FaceMAE to generate reconstructed dataset and train face recognition backbone on reconstructed dataset. The trained FaceMAE can be easily deployed on any masked face dataset without any extra training, which shows strong generalization of our proposed method.

In experiments, we verify that our FaceMAE outperforms the state-of-the-art synthetic face dataset generation methods by a large margin in terms of recognition accuracy in multiple large-scale face datasets. As shown in Fig. \ref{fig:motivation}, when training on reconstructed images from 75\% masked faces of CASIA-WebFace, FaceMAE consistently \textbf{reduces at least 50\% error rate} than the runner-up method (SynFace) on LFW, CFP-FP, and AgeDB datasets. It means better utility of the privacy-preserving face images generated by our method. We implement real-to-synthetic face retrieval experiments namely membership inference, and the results show that our FaceMAE makes membership inference significantly hard.

We summarize our contributions as:
\begin{itemize}
    \item {We propose a new paradigm for privacy-preserving face recognition, named FaceMAE, 
    that can be easily used on any face dataset to reduce the privacy leakage risk.
    }

    \item {Our FaceMAE jointly considers the privacy-persevering and recognition performance by adopting IRM module to minimize the distribution gap of original and reconstructed faces.}
    
    \item {Sufficient experiments verify that our FaceMAE outperforms the state-of-the-art methods, especially we \textbf{reduce at least 50\% error rate} than previous best method on LFW, CFP-FP, and AgeDB datasets. In addition, the risk of privacy leakage decreases around \textbf{20\%} in the face retrieval experiment.}

\end{itemize}

%% file: tex/relatedwork.tex
\section{Related Work}

\subsection{Face Dataset and Privacy}
Large-scale public face datasets \cite{parkhi2015deep, Guo2016MSCeleb1MAD, nech2017level, liu2018large, zhu2021webface260m} are typically collected by searching online images based on a name list of celebrities. Instead, those tech giants may obtain much more private face data \cite{taigman2015web, schroff2015facenet} for training their models. Both of them cause many privacy concerns, as they collect and store the original high-resolution personal images. 

Although differential privacy (DP) \cite{dwork2006calibrating, abadi2016deep} based methods have theoretical guarantees of privacy leakage and have been applied to easy datasets like MNIST \cite{lecun1998gradient}, it is not applicable to generate large-scale high-resolution face dataset, because of low utility of the generated samples \cite{xie2018differentially, cao2021don}. 
The recent works on generating privacy-preserving face dataset try to model the collect real faces with advanced GAN models and then synthesize fake face images with modifications \cite{wu2019privacy, li2019anonymousnet, mirjalili2020privacynet, bortolato2020learning, qiu2021synface}. For example, \cite{qiu2021synface} propose to synthesize identity-ambiguous faces by mixing two label vectors.
Different from de-identification methods that aim to remove the identity of face images \cite{wu2019privacy}, we expect the privacy-preserving dataset can be used to train downstream identification models. 
Besides the utility problem, the above methods all have to access the raw face images of the target identities for training generative models. 

Encoder-decoder models are learned in ~\cite{you2021reversible, yang2022invertible} that can convert private faces into privacy-preserving images and then invert them back to original ones. Although our FaceMAE also has encoder-decoder structure, our method is different from them in three main aspects: 1) Their protected images are hardly useful for downstream tasks, while ours can be used for training downstream models directly. 2) Their protected images can be inverted to original ones, while ours cannot be inverted. 3) They have to access raw face images, while our method only needs to access desensitized face images.
An orthogonal technique to relieve face privacy concerns can be federated face recognition \cite{meng2022improving, bai2021federated, aggarwal2021fedface}, while it is still challenging to achieve comparable performance to that of centralized training.



\subsection{Membership Inference Attack}
Membership inference \cite{brickell2008cost, li2010slicing, shokri2017membership} is a fundamental privacy problem in machine learning applications and the attack against membership is extremely destructive for medical and finical applications. In the popular membership inference attack protocol, a predictor is learned to infer the membership of specific sample in the training set given access to the black-box model \cite{shokri2017membership, sablayrolles2019white}. The averaged prediction accuracy or some other metrics can be derived based on the input and output of shadow models \cite{sablayrolles2019white, carlini2021membership, rezaei2021difficulty}.

However, this protocol is designed for small-scale classification datasets and it is not applicable to large-scale face datasets with millions of identities. Note that the representation model trained on a set of face identities will include only the embedding module and the identity classifier will be dropped. Thus, it is not suitable to predict the existence of a training face image based on its output, i.e. feature, in face recognition task. As aforemotioned, we propose to measure the membership privacy leakage based on real-synthetic face image retrieval.

\subsection{Training Sample Synthesis}
Training deep neural networks with synthetic samples is a promising solution when collecting real training data is expensive or infeasible for privacy issues. As the most successful generative model, Generative Adversarial Networks (GANs) \cite{goodfellow2014generative, brock2018large} are leveraged or adapted to synthesize training images for downstream tasks. For example, \cite{antoniou2017data} use GANs to augment training samples for few-shot target domain. 
\cite{zhang2021datasetgan, yang2022learning} learn to synthesize pixel-level image-annotation pairs, in order to minimize the manual annotation efforts. However, it is known that the synthetic and real data would have serious domain gap, and thus models trained on synthetic data perform badly on real testing data \cite{ravuri2019seeing, zhao2022synthesizing}.
To narrow the domain gap between synthetic and real face images, \cite{qiu2021synface} propose to mixup synthetic and real faces. \cite{zhao2022synthesizing} learn informative latent vectors of a pre-trained GAN model corresponding to informative synthetic training images by explicitly matching synthetic and real data distribution in many embedding spaces. Instead, we learn FaceMAE which can convert the largely masked faces to informative training samples for training high-performance recognition models.

%% file: tex/method.tex
\section{Proposed Method}
In this section, we first overview the pipeline of FaceMAE. Then, we briefly revisit the masked autoencoders (MAE) \cite{he2021masked} and formally introduce masked face reconstruction. After that, a carefully designed instance relation matching (IRM) module is presented. {Finally, we give the measurement of the membership privacy leakage of face datasets based on face image retrieval.}

\subsection{Overview of FaceMAE}
\begin{figure}[tp]
    \centering
    \includegraphics[width=\columnwidth]{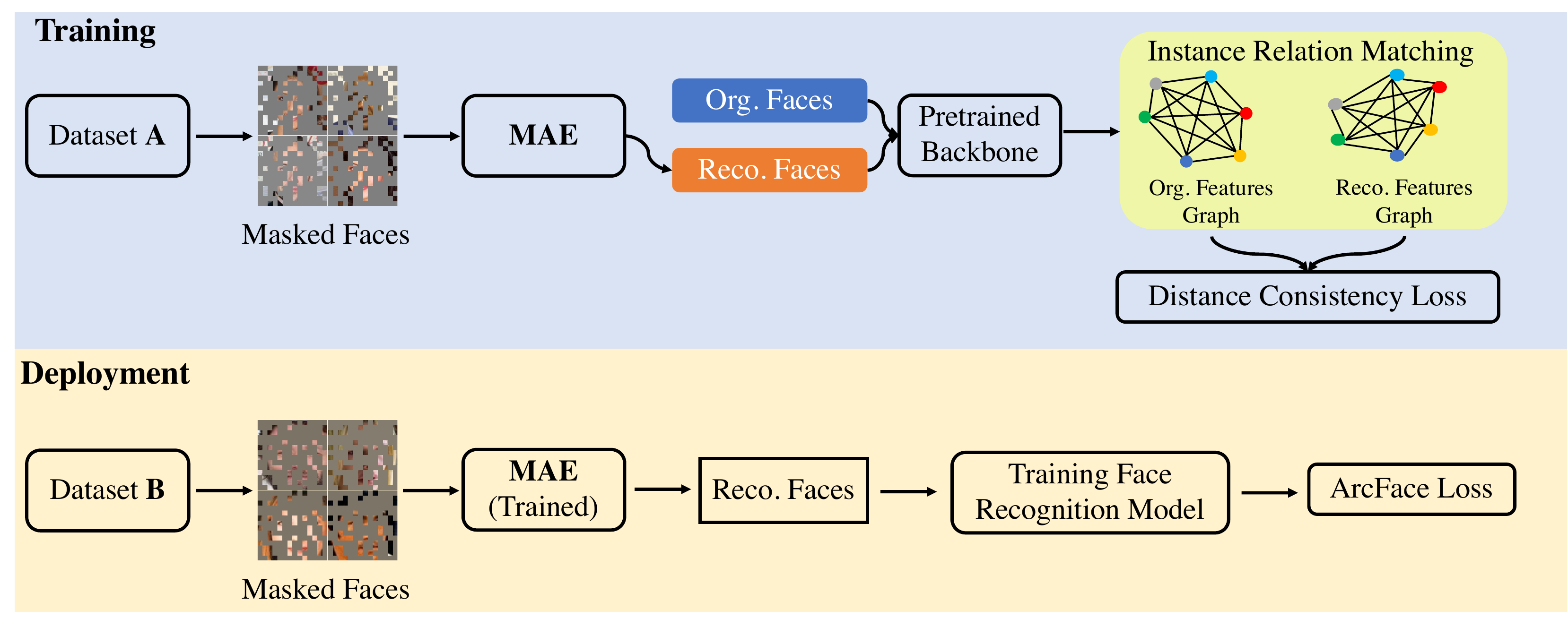}
    \caption{Illustration of the pipeline of FaceMAE.  Org. and Reco. denote original and reconstructed, respectively. In training stage, FaceMAE learns to generate reconstructed faces from masked faces. Instance relation matching (IRM) aims to ensure reconstructed faces that have similar distance graph with original ones in feature space. Distance consistency loss is used to optimize FaceMAE. In deployment stage, we directly apply trained FaceMAE on unseen masked dataset to obtain reconstructed faces for face recognition training.}
    \label{fig:pipeline}
\end{figure}

An overview of the FaceMAE is illustrated in Fig. \ref{fig:pipeline}. FaceMAE includes two stages, named training and deployment.
In training stage, we first randomly mask the original faces with (by default) 75\% ratio from a public dataset.
Then, the masked faces are fed into FaceMAE to obtain the reconstructed ones.
Instead of minimizing the Mean Squared Error loss (MSE-Loss) between original and reconstructed faces in pixel space, we tailor the instance relation matching (IRM) module to minimize distribution gap between real faces and FaceMAE reconstructed ones in feature space.
In deployment stage, we apply the trained FaceMAE to reconstruct the masked faces from another dataset with unseen identities and assemble these reconstructed faces as a privacy-preserving dataset for face recognition tasks. 
Note that FaceMAE is trained on one dataset with raw images and then deployed to masked faces from unseen identities in order to construct the privacy-preserving dataset for these unseen identities.

\subsection{Masked Face Reconstruction}
\textbf{Revisiting Masked Autoencoders.}
Masked Autoencoders (MAE) \cite{he2021masked} is a simple but efficient self-supervised pretraining strategy for image classification and its downstream tasks, such as object detection, instance segmentation, and semantic segmentation.
MAE utilizes an asymmetric encoder-decoder architecture to reconstruct the unseen patches from masked image. To train MAE, the training images are randomly masked as the input, and the MSE-Loss between the original image and MAE reconstructed ones is minimized. With MAE pre-training, data-hungry models like ViT-Large/-Huge \cite{dosovitskiy2020image} can be trained well with improved generalization performance.

Thanks to MAE's image completion ability, we adapt it to reconstruct privacy-preserving faces from the masked ones.
Specifically, given $N$ raw face images $\mathcal{I} = \{I_{i}\}|_{i=1}^{|\mathcal{I}|}$, we resize them to 224 $\times$ 224 size.
The resized faces are divided into non-overlapping 16 $\times$ 16 patches. 
Then, we randomly select a subset (\textit{e.g.} 25\%) of patches and mask (\textit{i.e.}, remove) the rest. 
Each patch represents a \textit{token embedding}.
Only the visible patches are fed into transformer encoder $E$ to obtain visible tokens. 
After that, a shallow decoder $D$ is designed to reconstruct the input image from visible mask tokens. The whole reconstruction progress can be formulated as 
\begin{equation}
\hat{I} = D(E[\gamma( I)]), 
\end{equation}
where $\gamma(\cdot)$ represents the random mask operation.

\subsection{Instance Relation Matching}
We expect the reconstructed face images are informative as the training samples for downstream tasks rather than high visual fidelity.
The vanilla MAE adopts MSE-Loss to minimizes differences between reconstructed and original faces in pixel space for visual fidelity, which may ignore the distribution gap between original faces and reconstructed ones.
In order to make reconstructed faces more informative as training samples, we design the instance relation matching (IRM) module to learn FaceMAE.
The IRM consists of two optimization components, named instance matching (IM) and relation matching (RM) losses.

With the raw and corresponding reconstructed images, we extract the features using an extra pre-trained embedding model $\phi(\cdot)$. We combine the features of the raw and reconstructed image set as two matrix $\bf{F} = [\phi(I_1), ..., \phi(I_N)\}]$ and $\hat{\bf{F}} = [\phi(\hat{I}_1), ..., \phi(\hat{I}_N)]$ respectively, where each row is a feature vector. We construct two graphs in the feature space with the data points as nodes and similarity as edges respectively. Then, the distribution consistency loss $\mathcal{L}_{dc}$ is calculated based on the matching of both nodes and edges of the two graphs, in other words, the instance matching and relation matching respectively. We formulate it as 
\begin{equation}
    \mathcal{L}_{dc} = \underbrace{\mathrm{\Delta}(\bf{F}, \hat{\bf{F}})}_{\mathrm{Instance \ Matching}} + \ \beta \underbrace{\mathrm{\Delta}(\langle \bf{F} \; , \; \bf{F}^\mathrm{T} \rangle, \langle \hat{\bf{F}} \; , \; \hat{\bf{F}}^\mathrm{T} \rangle)}_{\mathrm{Relation \ Matching}},
\end{equation}
where $\Delta$ measures the averaged pairwise Euclidean distance, $\langle \cdot \; , \; \cdot \rangle$ is the inner production, and $\beta$
is a trade-off coefficient between instance matching and relation matching.

\subsection{Privacy Leakage Risk}
\label{sec:method_privacy}
Our method is used to generate privacy-preserving face dataset. The potential attack to privacy is that the adversary may try to infer (retrieve) his/her interested identities in the face dataset by retrieving them with some other real faces. 
We measure the risk of this membership privacy leakage based on face image retrieval experiment. Specifically, we assume that the reconstructed set $\mathcal{S}=\{(\bm{s}_i, y_i)\}|_{i=1}^{|\mathcal{S}|}$ is released and attacker has a set of real faces $\mathcal{A}=\{(\bm{a}_i, z_i)\}|_{i=1}^{|\mathcal{A}|}$ of his/her interested identities. He/she can leverage well-trained  models to extract features and retrieve the set $\mathcal{R}$ based on $\mathcal{A}$. If the top-$K$ retrieved faces involve the ground-truth identity, we count it as one leakage event. Then, the averaged leakage risk $R(\mathcal{A}, \mathcal{S}, K)$ can be obtained by averaging multiple attacks:
\begin{equation}
\label{retr}
    R(\mathcal{A}, \mathcal{S}, K) = \frac{1}{\mathcal{A}}\Sigma_{i=1}^{|\mathcal{A}|} \1(z_i \in T(\bm{a}_i, \mathcal{S}, K))
\end{equation}
where $T(\bm{a}_i, \mathcal{S}, K)$ is a retrieval function that returns the label set of the $K$ reconstructed faces closest to the real query face $\bm{a}_i$ in embedding space. 

%% file: tex/experiments.tex
\section{Experiments}

\subsection{Datasets}
We first introduce the datasets used in training and deployment stages of FaceMAE and the three datasets for face verification.

\textbf{WebFace260M for FaceMAE Training.} By default, we use WebFace260M \cite{zhu2021webface260m} to train the FaceMAE. WebFace260M is the latest million-scale face benchmark, which is constructed for the research community. The high-quality training set of WebFace260M contains 42 millions faces of 2 millions identities. We use 10\% data for FaceMAE training.

\textbf{CASIA-WebFace for FaceMAE Deployment.} After training FaceMAE on WebFace260M, we deploy the trained FaceMAE on CASIA-WebFace \cite{yi2014learning}. 
The dataset includes 494, 414 face faces of 10, 575 real identities collected from the Internet.

\textbf{Face Verification Datasets.} We mainly validate our face recognition model on 3 face recognition benchmarks, including LFW~\cite{huang2008labeled}, CFP-FP\cite{7477558}, and AgeDB~\cite{moschoglou2017agedb}. 
 LFW is collected from the Internet which contains 13, 233 faces with 5, 749 IDs. CFP-FP collects 7, 000 celebrities' faces with 500 identities. Each identity contains 10 frontal view and 4 profile views. AgeDB contains 16, 488 images of various famous people, such as actors/actresses, writers, scientists, politicians, etc. Every image is annotated with respect to the identity, age and gender attribute. There exist a total of 568 distinct subjects.

 \subsection{Implementation Details}
Our FaceMAE is implemented with Pytorch \cite{paszke2019pytorch}. 
\textbf{Training Stage:} We follow the default hyper-parameters of original MAE.
The default patch size is 16, the size of input image is 224 $\times$ 224, the batch size is 256 per GPU, and mask ratio is 75\%.
For masked face reconstruction, we use eight A100 80G GPUs to train ViT-Base for 200 epochs (40 epochs for warm up).
The pretrained backbone (in Fig. \ref{fig:pipeline}) is ResNet-50 and its parameters is downloaded from InsightFace toolbox$\footnote{https://insightface.ai/}$.
The default $\beta$ is 1, the base learning rate is 1.5e-4, and the weight decay is 0.05.
\textbf{Deployment Stage:} We apply FaceMAE to reconstruct faces from masking CASIA-WebFace and train these reconstructed faces using ResNet-50 backbone from InsightFace. To make fair comparison, we set all hyper-parameters as same as InisghtFace. More details can be found in Supplementary Material.

\subsection{Comparison with State-of-the-art Methods}
\input{table/sota}
We compare FaceMAE with the state-of-the-art methods, including SynFace \cite{qiu2021synface}, FaceCrowd \cite{kou2022can} and ArcFace \cite{deng2019arcface}.
To investigate the effect of our proposed method, we also compare FaceMAE with vanilla MAE \cite{he2021masked}.
The face verification performances on on LFW, CFP-FP, and AgeDB are shown in Tab. \ref{tab:sota}.
These evaluation datasets contain most of face verification cases, including the Internet faces verification, frontal and profile views faces verification, and cross-age faces verification.
ArcFace \cite{deng2019arcface} is a popular margin-based loss for face recognition and we train a ResNet-50 using ArcFace loss as upper-bond results in first row of Tab. \ref{tab:sota}.
SynFace utilizes Disco-GAN \cite{deng2020disentangled} to generate non-existing faces using two existing face identities and train ResNet-50 on these generated faces.
FaceCrowd chooses partial areas from existing faces, such as eyes and month corners, to train face recognition using ResNet-18.
As shown in Tab. \ref{tab:sota}, FaceMAE outperforms previous state-of-the-art methods on all the three evaluation datasets with a large margin. 
Specifically, the error rate is reduced \textbf{at least 50\%} compared to previously best method SynFace \cite{qiu2021synface}.
On the one hand, this demonstrates the effectiveness of our method that can reconstruct more discriminative faces for training.
On the other hand, these significant improvements also prove the proposed FaceMAE has better utility for privacy-preserving face recognition.


\subsection{Ablation Studies}
We perform extensive ablation studies to illustrate the effects of our method. For better evaluation, we conduct experiments where we train FaceMAE on 75\% masking WebFace260M dataset and deploy it on CASIA-WebFace (unless otherwise specified).

\textbf{Evaluation of MSE-Loss and distribution consistency loss $\mathcal{L}_{dc}$.} We first analyze the effect of $\mathcal{L}_{dc}$ of FaceMAE.
Note that, to better understand the effect of two components of $\mathcal{L}_{dc}$, we separately consider instance matching (IM) loss and relation matching (RM) loss. 
Experimental results are shown in Tab. \ref{tab:abl_module}. As one can see, the best results are achieved when jointly using IM and RM, which shows the compatibility of the two losses. 
We also investigate the effect of the MSE-Loss.
Applying either IM or RM individually also brings non-trivial improvement than using MSE-Loss. Adding MSE-Loss with $\mathcal{L}_{dc}$ cannot further improve the face verification performance, which implies the $\mathcal{L}_{dc}$ can provide sufficient supervision for reconstruction. We visualize the reconstructed faces using MSE-Loss and $\mathcal{L}_{dc}$ respectively in Supplementary Material.

\textbf{Exploring the mask ratio.} Higher mask ratio means the less information of face is accessible, so the privacy could be protected better. Therefore, in order to minimize the privacy concerns and keep the verification performance simultaneously, we explore the max threshold of mask ratio that can achieve comparable results as original faces. We first randomly mask WebFace260M with four different mask ratios (30\%, 50\%, 75\%, and 90\%). Then, we train the FaceMAE models on these masking datasets, respectively. Finally, we deploy these trained models to reconstruct faces from masking CASIA-WebFace and report the verification performances on evaluation datasets. As shown in Tab. \ref{tab:mask_ratio}, the performance degrades slightly when mask ratio increase from 30\% to 75\%. However, the verification accuracy drops obviously when the mask ratio is set to 90\%. It shows that masking too many patches of faces cannot guarantee the face recognition performance. Therefore, considering both performance and privacy, we set mask ratio as 75\% by default.

\input{table/abl_v2_non_resize}
\input{table/abl2}

\textbf{Evaluating the generalization on variant datasets.} By default, we train FaceMAE on WebFace260M and apply it to reconstruct faces from masking CASIA-WebFace. 
To investigate the generalization of training and applying FaceMAE on variant datasets, we exchange the usage of WebFace260M and CASIA-WebFace, \textit{i.e.} WebFace260M as deploying dataset while CASIA-WebFace as training dataset.
Another concern is that the identities of the two datasets may overlap. To eliminate the influence of overlapping identities in training and deploying, we use two different 10\% subsets of WebFace260M, namely W.$_{sub1}$ and W.$_{sub2}$, where W. is the abbreviation of WebFace260M.  
{We show the experimental results in Tab. \ref{tab:cross_dataset}. Several findings can be concluded in the following. First, deploying the trained FaceMAE on larger datasets can obtain better verification performances.
Second, our proposed FaceMAE has good generalization on variant datasets, no matter identity overlap exists or not.}

\textbf{Evaluating the generalization of cross-architecture.} Another key advantage    of our method is that the reconstructed faces can be used to train variant architectures, such as ResNet (R$\cdot$), MobileFaceNet (MBF) \cite{chen2018mobilefacenets} and Vision Transformer (ViT-S) \cite{dosovitskiy2020image}. We train reconstructed faces using R18, R100, MBF, and ViT-S. The experimental results can be found in Tab. \ref{tab:cross_arch}. The reconstructed faces are not sensitive to network architectures, which proves the good generalization of proposed FaceMAE. Using deeper network (R100) outperforms shallower network (R18) with 0.97\%, 2.6\%, and 1.38\% on LFW, CFP-FP, and AgeDB. {Therefore, it is feasible to utilize a light model to learn FaceMAE and apply to deeper model for boosting face recognition performance.}

\textbf{Exploring $\beta$.} We define $\beta$ as a trade-off hyper parameter between instance matching loss and relation matching loss. 
In order to evaluate the sensitiveness of $\beta$, we study the verification performances on evaluation datasets when setting $\beta$ to be 0.1, 0.5, 1.0, and 10 in Tab. \ref{tab:beta}. One can find that the performance increases monotonically when $\beta$ rises from 0.1 to 1. While continually increasing $\beta$ to 10, the performance degrades significantly. It shows that over emphasizing the importance of relation matching may lead to less discriminative face reconstruction.

\textbf{Evaluating several types of mask.} We apply different mask strategies to evaluate the robustness of FaceMAE and show the results in Tab. \ref{tab:facial_area}. The comparable results of the first and second line show that changing random seed has very little influence on our proposed method. Masking eye or mouth consistently degrades the face recognition performances on three evaluation datasets, which indicates these areas are more important for the performance of face recognition. Meanwhile, masking these areas also cause difficulties for face alignment.





\input{table/vis_retr}
\subsection{Analysis of privacy leakage risk}
As formulated in Sec. \ref{sec:method_privacy}, we measure the risk of membership privacy leakage based on the face retrieval between original and reconstructed face images.
The simulated retrieval experiments are explored on CASIA-WebFace.
We here set $K$=2 and show the results in Fig. \ref{fig:last_vis}. One can find that the retrieval performances of `Org. to MAE.' and `Org. to FaceMAE' retrieval degrade largely with the number of identity increasing, while the performance of `Org. to Org.' retrieval only degrades slightly. 
The comparison verifies that the difficulty of membership inference by retrieving the reconstructed faces in dataset with real faces is significantly increased by our method.
Fig. \ref{fig:3a} shows that `Org. to MAE' retrieval is challenging compared to `Org. to Org.' baseline. This result verifies that applying MAE to face dataset is the key to reduce the privacy leakage.
`Org. to FaceMAE' retrieval accuracy is also obviously lower than `Org. to Org.' (illustrated in Fig. \ref{fig:3b}), around 20\% for 10000 IDs,  although it is higher than `Org. to MAE'.  The reason is that we trade-off the performance and privacy by introducing the instance relation matching module. Note that our FaceMAE achieves remarkable performance improvement in face recognition over MAE. 
Another possible privacy attack is that the adversary may imitate the masking process of MAE and retrieve the FaceMAE dataset with MAE processed images. Fig. \ref{fig:3c} shows that this kind of attack is unpromising which is much more challenging than `Org. to FaceMAE' retrieval.

%% file: table/sota.tex
\begin{table}[!t]
	\centering
	\small
\resizebox{\linewidth}{!}{
\begin{tabular}	{l |c|c|c|c|c }
		\toprule	 	
			\multirow{2}{*}{Method (Backbone)} &\multirow{2}{*}{Type} & \multirow{2}{*}{Training Dataset} & \multicolumn{3}{c}{Evaluation Datasets}\\
			\cmidrule{4-6}
			 &&& LFW & CFP-FP &AgeDB\\
			\midrule			
			ArcFace (R50)$^*$ \cite{deng2019arcface} &No Protection&CASIA-WebFace &99.35  &94.38&94.73 \\
			SynFace (R50)\cite{qiu2021synface} &GAN-Based&Syn-10K-50 &88.98 &76.23 &72.17\\
			FaceCrowd (R18)\cite{kou2022can} &Distortion&Partial Area of CelebA &83.58  &- &- \\
			ArcFace (R50)\cite{deng2019arcface} &Distortion&75\% Masking CASIA-WebFace &95.23  &79.02 &74.13 \\
			MAE + ArcFace (R50)\cite{deng2019arcface} &MAE-Based& Reconstructed CASIA-WebFace &97.62  &87.14 &85.68 \\
			
			\midrule
				\textit{FaceMAE} + ArcFace (R18) &MAE-Based& Reconstructed CASIA-WebFace &98.56  &89.31 &89.13 \\
			\textit{FaceMAE} + ArcFace (R50) &MAE-Based& Reconstructed CASIA-WebFace &\textbf{99.23}  &\textbf{90.80} &\textbf{90.25} \\
		
		\bottomrule
	\end{tabular}}
	\vspace{0.2em}
	\caption{Comparison with state-of-the-art methods using the metric of face verification accuracy (\%).
	$^*$ denotes the upper bond method that is reproduce by official InsighFace. Besides the upper bond results, the best entry of the rest results is marked in \textbf{bold}.}
	\label{tab:sota}	
\end{table}		

%% file: table/abl_v2_non_resize.tex
\begin{table}[h]
\begin{minipage}{0.32\linewidth}
    \centering
    \small
    \resizebox{\linewidth}{0.24\linewidth}{
    \begin{tabular}{ccc|ccc}
    \toprule
    
    
     \multicolumn{3}{c}{Operation} & \multicolumn{3}{|c}{Evaluation Datasets}\\
			\cmidrule{1-6}
			 MSE & IM &RM & LFW & CFP-FP &AgeDB\\
			\midrule
			\checkmark&&&97.62&86.14&85.68 \\
			 &\checkmark &&98.93&89.69&89.78\\
			&&\checkmark &98.92&87.10&86.68\\
			&\checkmark&\checkmark &\textbf{99.23}&\textbf{90.80}&\textbf{90.25}\\
			\checkmark&\checkmark&\checkmark&99.13&90.51&90.19\\
				\bottomrule
			 \end{tabular}
			 }
   \vspace{0.2em}
    \caption{Evaluation of MSE-Loss, IM and RM in the proposed \textit{FaceMAE}. \textbf{Bold entries} are best results.}
    
    \label{tab:abl_module}

\end{minipage}\hspace{3mm}\begin{minipage}{0.32\linewidth}
    \centering
    \small
    \resizebox{\linewidth}{0.24\linewidth}{
    \begin{tabular}{c|ccccc}
    \toprule
     \multirow{2}{*}{Mask Ratio} & \multicolumn{3}{|c}{Evaluation Datasets}\\
			\cmidrule{2-4}
			  & LFW & CFP-FP &AgeDB\\
			\midrule
			30\%&99.41&93.03&92.87\\
			50\%&99.33&91.98&91.56\\
			75\%&99.23&90.80&90.25\\
			90\%&67.33&57.38&54.01\\
			
			\bottomrule
			
    \end{tabular}}
     \vspace{0.2em}
    \caption{Evaluation of mask ratio of input faces for \textit{FaceMAE} training. All the mask patches are generated by random seed.}
    \label{tab:mask_ratio}

\end{minipage}\hspace{3mm}\begin{minipage}{0.32\linewidth}
    \centering
    \small
    \resizebox{\linewidth}{0.24\linewidth}{
    \begin{tabular}{c|c|cccc}
    \toprule
     \multirow{2}{*}{Train} & \multirow{2}{*}{Deploy} & \multicolumn{3}{c}{Evaluation Datasets}\\
			\cmidrule{3-5}
			 && LFW & CFP-FP &AgeDB\\
			\midrule
			
			W. & C. &99.23&90.80&90.25\\
			C. & W. &99.37&93.87&94.14\\
			W.$_{sub1}$ & W.$_{sub2}$ &99.65&94.00&94.36\\
			W.$_{sub2}$ & W.$_{sub1}$ &99.47&94.12&94.51\\
			
			
			\bottomrule
			
    \end{tabular}}
     \vspace{0.2em}
    \caption{Evaluation of the generalization of \textit{FaceMAE}. W. and C. represent WebFace260M and CASIA-WebFace.}
    \label{tab:cross_dataset}

\end{minipage}
\end{table}


%% file: table/abl2.tex
\begin{table}[h]
\begin{minipage}{0.32\linewidth}
    \centering
    \small
       \resizebox{\linewidth}{0.24\linewidth}{
    \begin{tabular}{c|ccccc}
    \toprule
     \multirow{2}{*}{Arch.} & \multicolumn{3}{|c}{Evaluation Datasets}\\
			\cmidrule{2-4}
			  & \;\; LFW \;\;& \;\;CFP-FP \;\;&\;\;AgeDB\;\;\\
			\midrule
			R18&98.56&89.31&89.13\\
			R100&99.53&91.91&90.51\\
			MBF&98.40&88.94&88.56\\
			ViT-S&98.45&90.12&90.18\\
			
			\bottomrule
			
    \end{tabular}}
     \vspace{0.2em}
    \caption{Exploration of cross-architecture generalization of FaceMAE. R$\cdot$, MBF and ViT-S denote ResNet, MobileFaceNet, and Vision Transformer.}
    \label{tab:cross_arch}

\end{minipage}\hspace{3mm}\begin{minipage}{0.32\linewidth}
    \centering
    \small
    \resizebox{\linewidth}{0.24\linewidth}
    {
    \begin{tabular}{c|ccc}
    \toprule
     \multirow{2}{*}{$\beta$} & \multicolumn{3}{|c}{Evaluation Datasets}\\
			\cmidrule{2-4}
			  \;\;\;\;& \;\;\;\;LFW \;\;& \;\;CFP-FP \;\;& \;\;AgeDB\;\;\;\;\\
			\midrule
			0.1&98.94&88.21&87.13\\
			0.5&99.02&89.01&88.96\\
			1.0&99.23&90.80&90.25\\
			10&98.97&87.65&86.96\\
			
			\bottomrule
			
    \end{tabular}}
     \vspace{0.2em}
    \caption{Exploration of different $\beta$ from 0.1 to 10. $\beta$ is a trade-off parameter between instance matching loss and relation matching loss.}
    \label{tab:beta}

\end{minipage}\hspace{3mm}\begin{minipage}{0.32\linewidth}
    \centering
    \small
    \resizebox{\linewidth}{0.24\linewidth}{
    \begin{tabular}{c|ccccc}
    \toprule
     \multirow{2}{*}{Mask Area} & \multicolumn{3}{|c}{Evaluation Datasets}\\
			\cmidrule{2-4}
			  & LFW & CFP-FP &AgeDB\\
			\midrule
			Seed 0&99.23&90.80&90.25\\
			Seed 1&99.10&90.67&90.21\\
			Eye&98.45&89.98&89.76\\
			Mouth&98.77&90.14&89.93\\
			
			\bottomrule
			
			
			
    \end{tabular}
    }
     \vspace{0.2em}
    \caption{Exploring which facial area is more related to the face recognition performance. The first and second rows use different random seeds for masking.}

    \label{tab:facial_area}

\end{minipage}
\end{table}


%% file: table/vis_retr.tex
\begin{figure*}[htp]

    \centering
    
    \subfigure[]
    {\label{fig:3a}{\includegraphics[width=0.32\textwidth]{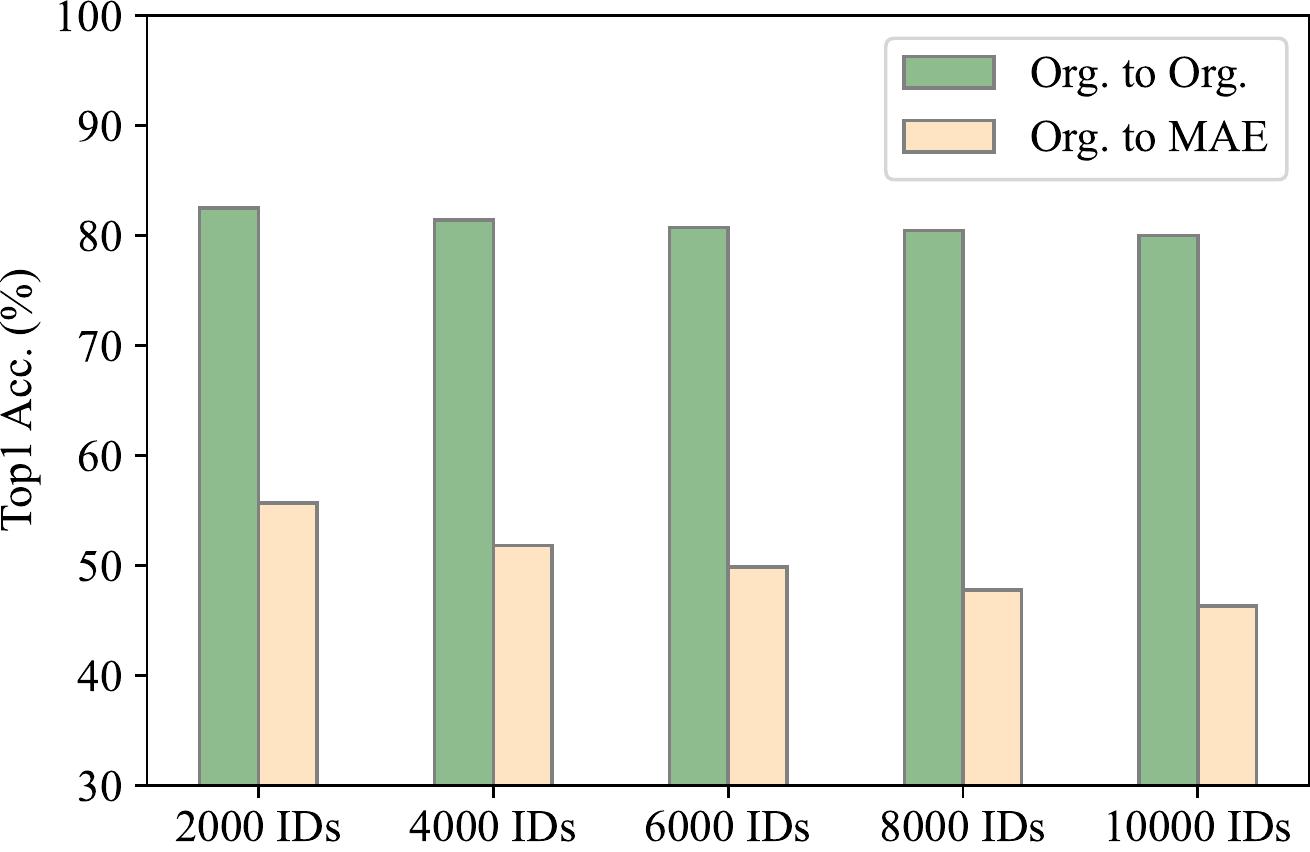}}}
    \subfigure[]
    {\label{fig:3b}{\includegraphics[width=0.32\textwidth]{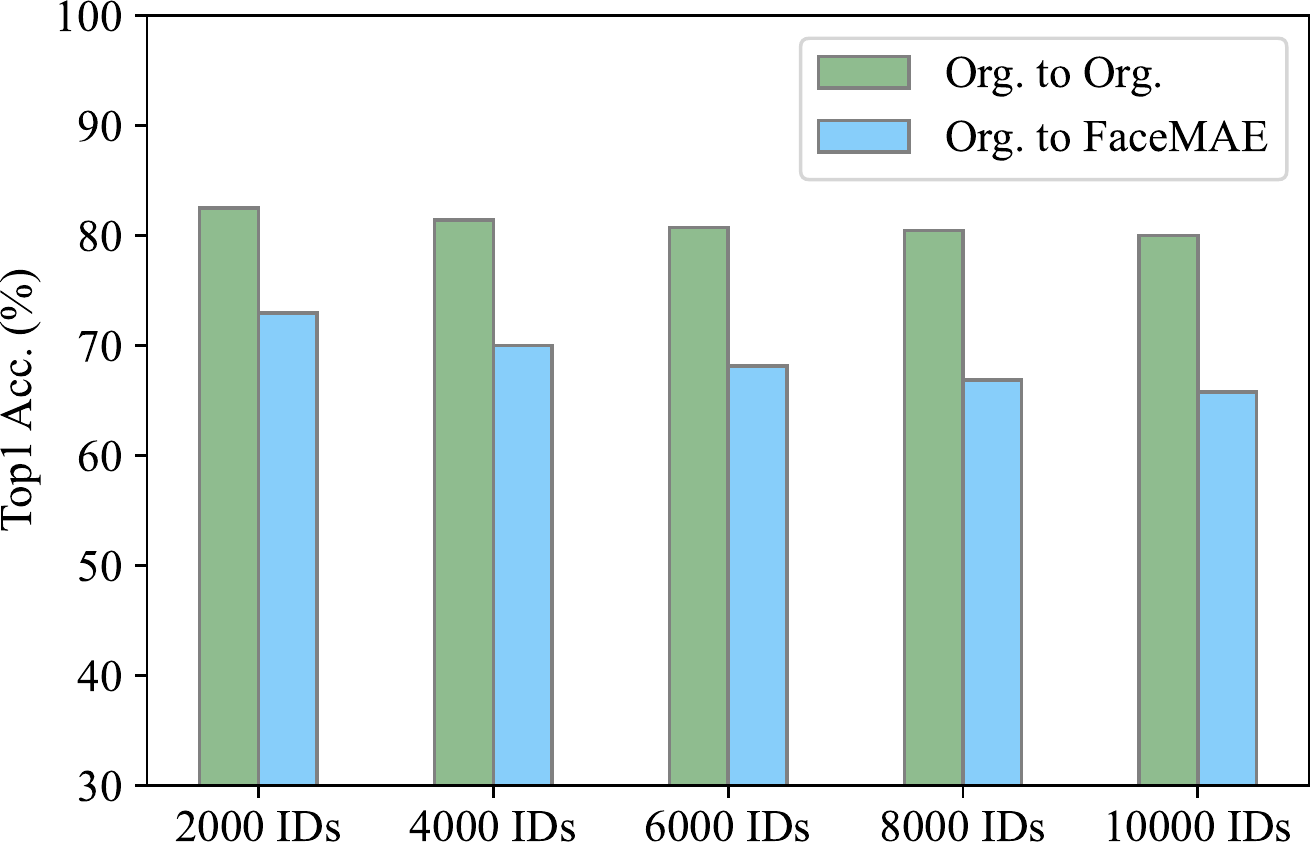}}}
    \subfigure[]
    {\label{fig:3c}{\includegraphics[width=0.32\textwidth]{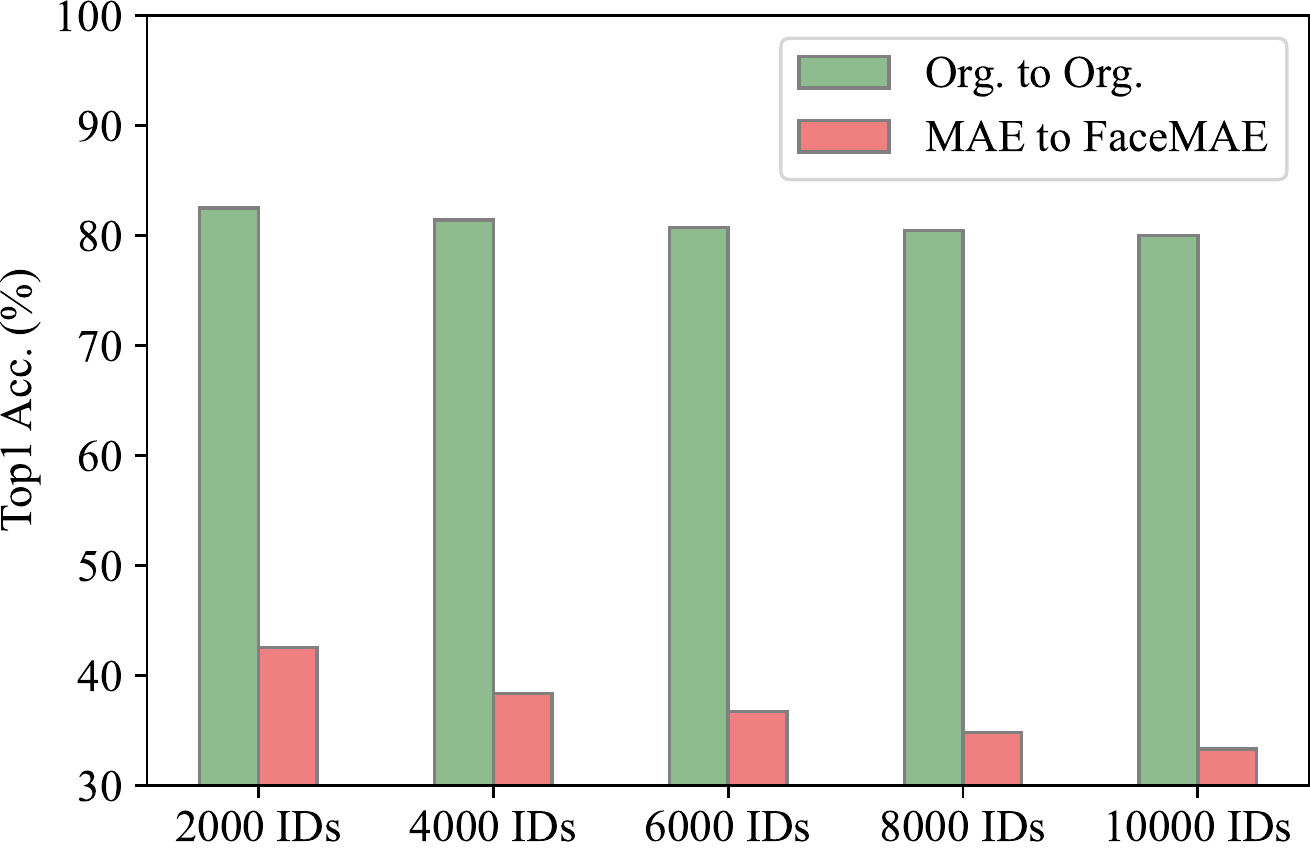}}}
\caption{(a), (b), and (c) compare the retrieval performances on Original (Org.), MAE reconstructed and FaceMAE reconstructed datasets. The green bar (`Org. to Org.') represents using images from original dataset to retrieve each other. Other colors denote the retrieval experiments are completed among different settings. 
\label{fig:last_vis}}
\end{figure*}

%% file: tex/conclusion.tex
\section{Conclusion \& Discussion}
This paper deals with an urgent and challenging problem of privacy-preserving face recognition. FaceMAE is proposed to generate synthetic samples that reduce the privacy leakage and maintain recognition performance simultaneously. Instance relation matching module instead of MSE loss in vanilla MAE is designed to enable generated samples to train deep models effectively. The experiments verify that the proposed MAE surpasses the runner-up method by reducing at least 50\% recognition error on popular face dataset. The privacy leakage risk decreases around 20\% when our FaceMAE is applied. 

\paragraph{Discussion.} As a new framework for protecting the face membership privacy, our FaceMAE still has several limitations. First, our FaceMAE can reduce the membership privacy leakage of a face dataset rather than completely solve this problem. More future work is needed to advance large-scale privacy-preserving face recognition. Second, our method doesn't defend the inversion attack on model or training sample, which is out of the scope of this paper. Fortunately, those defence methods can be easily applied together with our method. 